\documentclass[letterpaper, 10 pt, conference]{ieeeconf}  
\IEEEoverridecommandlockouts                              
\usepackage[bookmarks=true]{hyperref}
\usepackage{cite}
\usepackage{amsmath,amssymb,amsfonts}
\usepackage{graphicx}
\usepackage{gensymb}
\usepackage{textcomp}
\usepackage{xcolor}
\usepackage{placeins}
\usepackage{titlesec}
\usepackage{comment}
\usepackage{wrapfig}
\usepackage{placeins}
\usepackage[caption=false, font=footnotesize]{subfig}
\usepackage{multirow}
\usepackage{makecell}
\usepackage{mathtools}
\usepackage[nolist,nohyperlinks]{acronym}
\usepackage{colortbl}
\usepackage{pdfpages}
\usepackage{algorithm}
\usepackage{algpseudocode} 
\usepackage[skip=2pt,font=footnotesize]{caption}
\let\labelindent\relax
\usepackage{enumitem}
\setlist{nosep}

\usepackage{amsmath}

\newlist{todolist}{itemize}{2}
\setlist[todolist]{label=$\square$}
\usepackage{pifont}

\usepackage[capitalise]{cleveref} 

\crefname{algorithm}{algorithm}{algorithms} 
\Crefname{algorithm}{Algorithm}{Algorithms} 

\usepackage{needspace}
\usepackage{mwe}

\usepackage{tikz}

\newcommand{\blackcircle}{%
    \raisebox{-0.05ex}{\tikz\fill[black] (0.08,0.08) circle (0.08);}%
}

\newcommand{\blackstar}{%
    \raisebox{-0.05ex}{\tikz\fill[black] (0,0) -- (0.1,0.3) -- (0.2,0) -- (-0.05,0.18) -- (0.25,0.18) -- cycle;}%
}

\acrodef{qp}[QP]{Quadratic Programming}
\acrodef{fov}[FOV]{field of view}
\acrodef{ppo}[PPO]{Proximal Policy Optimization}
\acrodef{pomdp}[POMDP]{partially observable Markov decision process}
\acrodef{ttc}[TTC]{time-to-collision}
\acrodef{sota}[SOTA]{state of the art}
\acrodef{cr}[CR]{crash rate}
\acrodef{sr}[SR]{success rate}
\acrodef{tof}[ToF]{Time-of-Flight}
\acrodef{ood}[OOD]{out-of-distribution}
\acrodef{rio}[RIO]{Radar Inertial Odometry}

\title{\LARGE \bf
Sensor-Layout-Agnostic Navigation via Geometric Observation Canonicalization
}

\author{Welf Rehberg and Kostas Alexis%
\thanks{The authors are with the Norwegian Institute of Science and Technology, Norway. \{welf.rehberg, konstantinos.alexis\}@ntnu.no}
\thanks{This work was supported by the Horizon Europe Grant Agreement No. 101119774, the Research
Council of Norway under Award NO-357451 and the NVIDIA Academic Grant Program using NVIDIA RTX PRO 6000 Blackwell Max-Q and A100 GPU-Hours on Saturn Cloud.}
}

\titlespacing*{\section}
  {0pt}   
  {1.5ex plus 1ex minus .2ex}  
  {0.5ex plus .2ex}            

\titlespacing*{\subsection}
  {0pt}
  {1ex plus .2ex minus .2ex}
  {0.3ex plus .2ex}

\allowdisplaybreaks

\begin{document}

\maketitle
\thispagestyle{empty}
\pagestyle{empty}

\begin{abstract}
Existing visual navigation policies are inherently bound to fixed camera configurations, creating a fundamental barrier to zero-shot deployment across heterogeneous robot sensor layouts. To overcome this limitation, we present an embodiment-informed navigation policy capable of generalizing across diverse depth sensor configurations on a specific aerial platform. Instead of implicitly learning spatial alignments, our approach explicitly unprojects depth measurements from arbitrary depth sensor payloads, varying in sensor count, mounting extrinsics, and intrinsics, into a shared robot-centric frame, stitching them into a unified spherical range image and a binary validity mask. This mask allows the downstream policy to explicitly distinguish covered space from unobserved blind spots. Trained via reinforcement learning with aggressive camera randomization, our policy generalizes zero-shot to unseen layouts featuring up to seven cameras, scaling success rates from $78\mathrm{\%}$ to $95\mathrm{\%}$ as total spatial sensing coverage increases. Finally, real-world flight trials on a physical quadrotor, conducted in an obstacle-filled corridor and an outdoor forest, validate the policy's zero-shot transfer across camera configurations and its resilience to sudden online sensor dropouts.

\end{abstract}
\section{Introduction}
\FloatBarrier
\begin{figure}[t]
    \centering
    \includegraphics[width=0.8\linewidth]{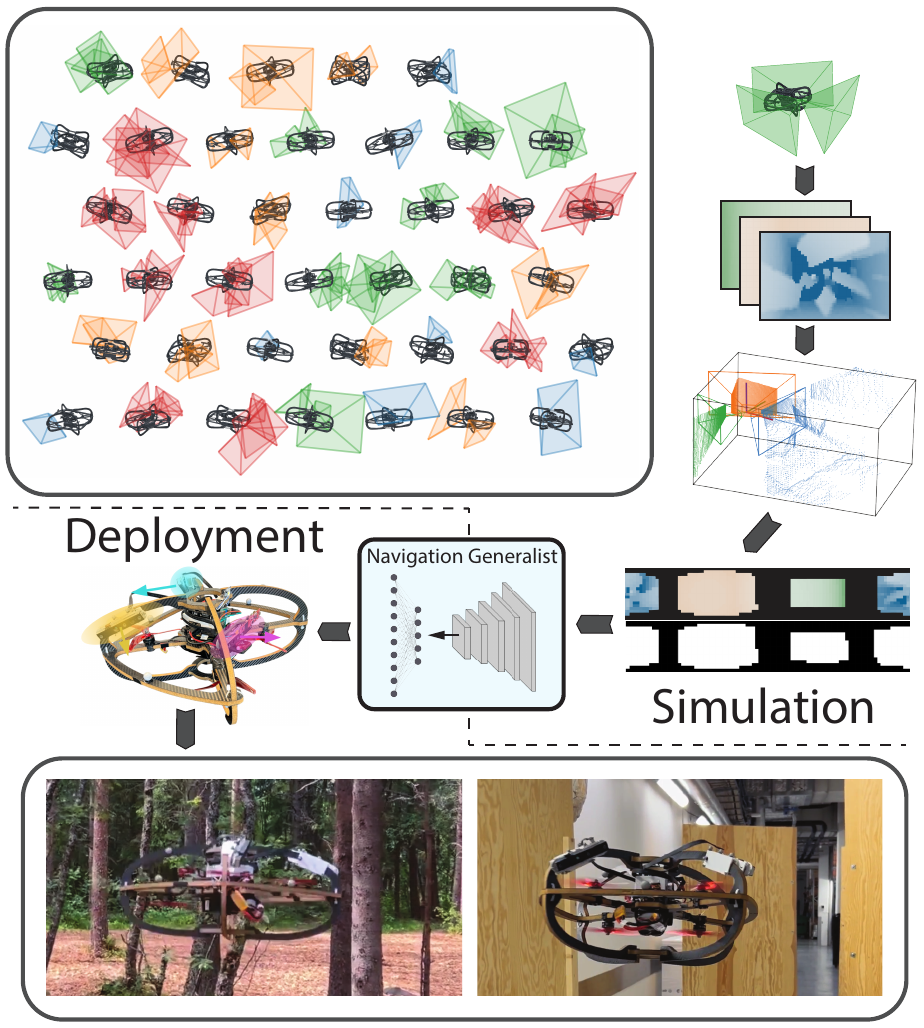}
    \caption{Overview of the navigation generalist method.}
    \label{fig:beauty_figure}
\end{figure}

Deploying diverse autonomous robots across unstructured environments often requires custom tuning or re-training policies whenever sensor hardware changes. Consequently, zero-shot generalist policies are essential to eliminate per-platform engineering and enable seamless deployment across diverse, evolving robot fleets. To navigate reliably, a generalist policy must ingest raw sensor information observations and map them directly to control commands across highly diverse embodiments. A major barrier to this generalization is that navigation policies are historically bound to specific camera configurations and vehicle frames \cite{loquercioLearning2021, florenceIntegrated2020, hengAutonomous2011}. This tight coupling means even minor adjustments to the sensor payload, such as varying mounting heights, altering pitch angles, or integrating auxiliary cameras, render the policy unusable, traditionally demanding expensive network retraining.

Recent research has progressed significantly on generalizing policies to robots with different dynamic responses \cite{zhang2023learning,raptor,konstantaropoulos2026embodiment} and kinematic chains \cite{bohlinger2025policyrunallendtoend,luo2025gcntgraphbasedtransformerpolicies}. However, existing visual generalists either suffer from high computational latency or rely on simplified geometric assumptions, thereby preventing lightweight, real-time 3D navigation across heterogeneous camera layouts.
In this work, we address the problem of reactive local collision avoidance for a broad family of multi-sensor depth camera setups on a fixed dynamical system. Rather than forcing the neural network to implicitly associate visual coordinates with motion primitives, we inject a strong geometric prior into the network's perception pipeline. We unproject exteroceptive depth measurements from a variable number of arbitrary sensors and map them into a shared, unified egocentric spherical coordinate system, yielding a range image and an associated binary validity mask expressed in the robot's base coordinate frame. Canonicalization projects all sensor measurements into a common coordinate system, removing the need for the policy to implicitly learn sensor-to-body frame alignments. An overview of the method is shown in \cref{fig:beauty_figure}.
Through extensive simulated evaluations and real-world quadrotor deployments, we demonstrate that our embodiment-informed policy achieves robust, zero-shot transfer across highly heterogeneous exteroceptive configurations and environments and is robust to sensor dropouts on systems with redundant sensor configurations. Specifically, the contributions of this paper are four-fold:
\begin{itemize}
    \item We introduce an exteroceptive canonicalization pipeline that maps arbitrary depth feeds to a unified, egocentric spherical range tensor and a binary validity mask, enabling generalization to sensor count, mounting extrinsics, and sensor intrinsics.
    \item We present a reactive navigation policy trained via reinforcement learning under aggressive sensor extrinsics and intrinsics randomization, eliminating the need to implicitly learn sensor-body-coordinate alignments.
    \item We demonstrate zero-shot generalization in simulation to \ac{ood} layouts featuring up to seven cameras, achieving navigation performance comparable to specialized, layout-specific policies.
    \item We validate our approach on a physical quadrotor across different sensor configurations navigating cluttered indoor corridors and dense forests, demonstrating robust real-world zero-shot transfer and resilience to sudden online sensor dropouts.
\end{itemize}
The remainder of this paper is organized as follows. \cref{sec:related_work} reviews prior work on generalist navigation and cross-embodiment learning. \cref{sec:prob_form} introduces the problem formulation. \cref{sec:method} details the system family considered in this work, the geometric observation canonicalization method, our architecture choices, and our training setup. \cref{sec:experiments} presents our ablation studies, simulation benchmarks, and real-world flight deployments. Finally, \cref{sec:conclusion} concludes the paper and discusses future research directions.

\section{Related Work}
\label{sec:related_work}
Within the domain of cross-embodiment control, existing research predominantly focuses on adapting low-level motor or joint commands to variations in physical dynamics, such as mass, inertia, and joint kinematics \cite{raptor, zhang2023learning, bohlinger2025policyrunallendtoend, konstantaropoulos2026embodiment, luo2025gcntgraphbasedtransformerpolicies}. However, this body of work leaves the challenge of exteroceptive perception geometry largely unaddressed, assuming static or platform-specific sensor arrangements. 
Recently, extending generalization capabilities to visual navigation has gained significant momentum. Frameworks such as GNM \cite{ShahGNM2023}, ViNT \cite{shah2023vint}, and X-Nav \cite{Wang2026XNav} enable generalized navigation across diverse physical embodiments, spanning mobile robots, quadrupeds, and aerial vehicles. Despite their success, these architectures operate under a restrictive assumption: they assume a fixed nominal sensor payload per platform, most commonly a single forward-facing camera. While these models implicitly absorb minor camera height or pitch variations that are present across large training datasets, they lack explicit mechanisms to handle multi-camera setups, arbitrary $3$D mounting extrinsics, or variable sensor counts.
While X-Nav \cite{Wang2026XNav} targets hardware-agnosticism by deterministically projecting depth images into a $1$D laser scan format, their training pipeline and evaluation are restricted to a single nominal camera configuration with a $90^\circ$ \ac{fov}. Consequently, the policy is never exposed during training to vastly different camera poses, different camera intrinsics, multi-camera setups, or unobserved spatial gaps. Furthermore, reducing depth observations to a $1$D horizontal representation fundamentally restricts the system to a $2$D plane, precluding its application to $3$D aerial flight where elevation-aware obstacle avoidance is essential. 

Generalist vision-language-action (VLA) navigators \cite{eftekhar2025the, zhang2026embodied} incorporate multi-view visual feeds by scaling up high-capacity visual transformers to associate RGB tokens with high-level semantic subgoals. However, these models target low-frequency planning rather than local collision avoidance and suffer from high computational latency (\cite{zhang2026embodied} requiring off-board GPU infrastructure), making them incapable of the high-frequency control loops needed for real-time onboard flight in cluttered environments. Additionally, the work in \cite{eftekhar2025the} severely limits the number of sensors (max $2$) and assumes that one sensor faces strictly forward, while \cite{zhang2026embodied} accommodates up to $8$ cameras but randomizes only mounting height and horizontal \ac{fov} continuously, selecting the underlying sensor layout from $6$ predefined discrete configurations. 

To overcome these limitations, we introduce a single generalist policy for real-time, onboard $3$D reactive collision avoidance across heterogeneous sensor layouts. Rather than relying on heavy transformer backbones or predefined sensor configurations, our lightweight model runs onboard embedded hardware, operating zero-shot across arbitrary sensor counts, extrinsics, and intrinsics. By unprojecting arbitrary depth feeds into an egocentric spherical range tensor paired with a binary validity mask, we decouple policy execution from viewpoint alignment. Trained across wide continuous variations in camera counts, $3$D mounting extrinsics and sensor intrinsics, our policy architecture learns to reason over visibility and obstacle proximity, achieving robust zero-shot \ac{ood} transfer and online fault tolerance on physical hardware.

\section{Problem Formulation}
\label{sec:prob_form}
We consider the problem of generalizing a single navigation policy across a family of systems characterized by their morphology vectors $\boldsymbol{\gamma}\in\Gamma$, where $\Gamma$ denotes the considered family of systems. We formulate this problem as a \ac{pomdp}, where the policy must  -for a single dynamical system- generalize across varying depth camera setups and incorporate observations from up to $n_\text{cam}$ arbitrarily placed and oriented depth sensors while navigating tight, cluttered environments. The morphology vector is defined by the corresponding sensor extrinsics ${}^{\mathcal{B}}\boldsymbol{T}_{\mathcal{C}_j}$ and intrinsics $\boldsymbol{K}_j$ as
\vspace{-0.2cm}
\begin{equation}
    \boldsymbol{\gamma} = [ \text{vec}([{}^{\mathcal{B}}\boldsymbol{T}_{\mathcal{C}_1},...,{}^{\mathcal{B}}\boldsymbol{T}_{\mathcal{C}_{n_\text{cam}}}]),\text{vec}([\boldsymbol{K}_1,...,\boldsymbol{K}_{n_\text{cam}}])].
\end{equation}
Here, $\mathcal{B}$ denotes the body frame of the robot and $\mathcal{C}_j$ denotes the coordinate frame of sensor $j$ and $\text{vec}(\cdot)$ denotes the vectorize operation. 
Given a 3D start position $\boldsymbol{p}_s \in \mathbb{R}^{3}$ and a goal position $\boldsymbol{p}_g \in \mathbb{R}^{3}$, both expressed in the inertial frame $\mathcal{I}$, the objective is to generate a sequence of actions $\boldsymbol{a}_t=[a_{t,x},a_{t,y},a_{t,z},a_{t,\omega}]$ with the first three components being linear reference velocities and $a_{t,\omega}$ being the reference yaw rate, all expressed in the body frame $\mathcal{B}$. At each timestep $t$, the policy has access only to (a)~a proprioceptive state estimate
$\boldsymbol{x}_t = [\boldsymbol{g}_t^{\!\top},\, \xi_t,\, \eta_t,\, \boldsymbol{v}_t^{\!\top},\, \boldsymbol{\omega}_t^{\!\top}]^{\!\top} \in \mathbb{R}^{12}$,
where $\boldsymbol{g}_t = [\hat{\boldsymbol{d}}_t^{\!\top},\, d_t]^{\!\top} \in \mathbb{R}^{4}$ concatenates the unit direction
$\hat{\boldsymbol{d}}_t \in \mathbb{R}^{3}$ from the robot to the goal and the Euclidean goal distance $d_t \in \mathbb{R}$,
both expressed in $\mathcal{B}$;
$\xi_t, \eta_t \in \mathbb{R}$ denote the roll and pitch angles;
and $\boldsymbol{v}_t, \boldsymbol{\omega}_t \in \mathbb{R}^{3}$ are the linear and angular velocities expressed in $\mathcal{B}$;
(b)~the previous action $\boldsymbol{a}_{t-1}$;
 and (c) the depth observations $d_j(u,v)$ (axial), with $j \in {1,\ldots,n_\text{cam}}$, together with the sensor extrinsics ${}^{\mathcal{B}}\boldsymbol{T}_{\mathcal{C}_j}$ and intrinsics $\boldsymbol{K}_j$.
\section{Method}
\label{sec:method}
\subsection{System Family}
\label{sec:system_representation}

\begin{figure}[t]
\centering
\begin{minipage}[c]{0.48\linewidth}
\centering
\scriptsize
\captionof{table}{Camera extrinsic and intrinsic randomization ranges.}
\label{tab:camera_randomization}
\begin{tabular}{lc}
\hline
Parameter & Range \\
\hline
Number of cameras & $\{1,2,3,4\}$\\
\hline
\multicolumn{2}{l}{\textbf{Extrinsics}} \\
Position ($x,y$)  & $[-0.2\mathrm{m},\,0.2\mathrm{m}]\,$ \\
Position ($z$)  & $[-0.1\mathrm{m},\,0.1\mathrm{m}]\,$ \\
Roll  & $[-90^\circ,\,90^\circ]$ \\
Pitch  & $[-45^\circ,\,45^\circ]$ \\
Yaw  & $[-180^\circ,\,180^\circ]$ \\
\hline
\multicolumn{2}{l}{\textbf{Intrinsics}} \\
Focal length & $[1\mathrm{mm},\,3\mathrm{mm}]$ \\
Horizontal aperture & $[2.8\mathrm{mm},\,4.4\mathrm{mm}]$ \\
Aperture aspect ratio & $[0.75,\,2.0]$ \\
Principal point & image center (fixed) \\
\hline
\end{tabular}
\end{minipage}
\hfill 
\begin{minipage}[c]{0.4\linewidth}
\centering
\includegraphics[width=\linewidth]{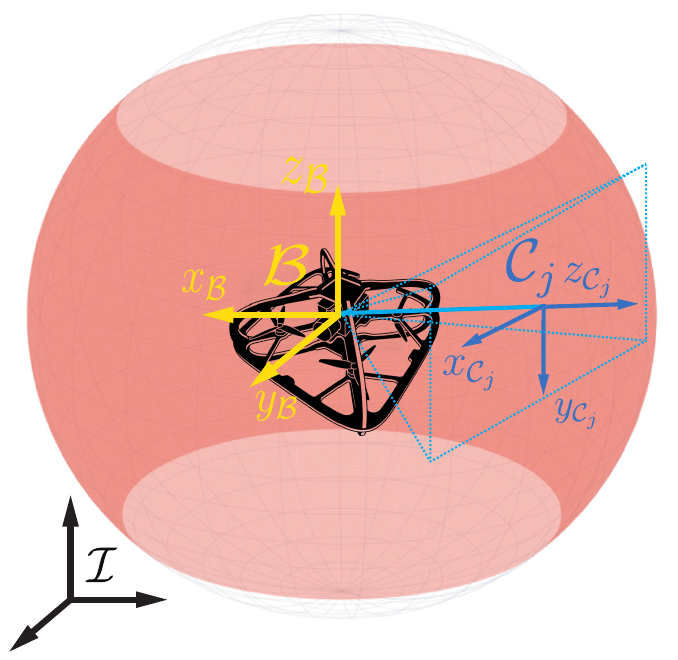}
\caption{Feasible area of the principal axis for configurations considered in this work.}
\label{fig:random_scheme}
\end{minipage}
\end{figure}

The family of systems to which the policy is expected to generalize to is created by sampling configurations in the following manner. For a fixed multirotor system, the number of cameras is uniformly sampled from the set $\{1,...,n_\text{cam}\}$ ($n_\text{cam}=4$ in this work). The camera extrinsics are randomized as follows. The position of camera $j$ in the body frame, ${}^\mathcal{B}\boldsymbol{t}_{\mathcal{C}_j}$, is sampled uniformly from a box enclosing the full robot, with bounds detailed in \cref{tab:camera_randomization}. The orientation of camera $j$ in the body frame, ${}^{\mathcal{B}}\boldsymbol{R}_{\mathcal{C}_j}$, is restricted to configurations favorable for navigation, excluding cameras oriented fully upward or downward as illustrated in \cref{fig:random_scheme} and uniformly sampled from the ranges detailed in \cref{tab:camera_randomization}. The camera intrinsics $\boldsymbol{K}_j$ are randomized by uniformly sampling the focal length $f$, the horizontal aperture $a_h$, and the aperture aspect ratio $\rho_a= a_h/a_v$ from the ranges detailed in \cref{tab:camera_randomization}. Where $a_v$ is the vertical aperture. The resulting \ac{fov} ranges from $[50^\circ,131^\circ]$ in azimuth and $[26^\circ,142^\circ]$ in elevation. In practice, the intrinsic matrix $\boldsymbol{K}_j$ is obtained from these randomized optical parameters. Given an image resolution of $W \times H$, the focal lengths in pixel coordinates are computed as
\vspace{-0.2cm}
\begin{equation}
f_x = W\frac{f}{a_h}, \qquad
f_y = H\frac{f}{a_v}, \qquad
a_v = \frac{a_h}{\rho_a}.
\end{equation}

We randomize the aspect ratio rather than the vertical aperture to uniformly cover narrow and wide \ac{fov}s. The principal point is not randomized and is fixed at the image center, $(c_x,c_y)=(W/2,H/2)$, since any known spatial offsets can be fully compensated for during preprocessing of the depth scan. Note that the image resolution during training is fixed, but as described in \cref{sec:obs_canon} resampling the real image to the target resolution used in training lets us deploy cameras with varying resolutions. The system dynamics are not randomized and instead, a nominal system model is assumed. The configuration-defining extrinsics and intrinsics matrices are therefore
\vspace{-0.2cm}
\begin{equation}
{}^\mathcal{B}\boldsymbol{T}_{\mathcal{C}j}
=
\left[
\begin{array}{cc}
{}^\mathcal{B}\boldsymbol{R}_{\mathcal{C}j} &
{}^\mathcal{B}\boldsymbol{t}_{\mathcal{C}_j} \\
\boldsymbol{0}^T & 1
\end{array}
\right],
\boldsymbol{K}_j =
\left[
\begin{array}{ccc}
f_x & 0 & c_x \\
0 & f_y & c_y \\
0 & 0 & 1
\end{array}
\right].
\end{equation}
\vspace{-0.1cm}
\subsection{Observation Canonicalization}
\label{sec:obs_canon}

\subsubsection{Canonicalized Spherical Range Image}
To enable the policy to operate on sensor readings obtained from arbitrarily placed cameras, we transform the obtained depth scans $d_j(u,v)$ ($j \in \{1,..,n_\text{cam}\}$) into patches in a spherical range image $r(i_{\phi}, i_{\theta})$. 
Each depth pixel $(u,v)$ is mapped to a point in the robot base frame $\mathcal{B}$ using the corresponding sensor intrinsic $\boldsymbol{K}_j$ and extrinsic ${}^\mathcal{B}\boldsymbol{T}_{\mathcal{C}j}$:
\begin{equation}
    \begin{bmatrix} \boldsymbol{p}_\mathcal{B} \\ 1 \end{bmatrix} = {}^\mathcal{B}\boldsymbol{T}_{\mathcal{C}_j} \begin{bmatrix} \boldsymbol{K}_j^{-1} \begin{bmatrix} u & v & 1 \end{bmatrix}^T d_j(u,v) \\[0.5ex] 1 \end{bmatrix}.
\end{equation}
Note that for depth sensors that cannot be modeled using a pinhole camera model, the mapping must be adapted accordingly (outside the scope of this work). Subsequently, the 3D point cloud in Cartesian robot coordinates $\boldsymbol{p}_\mathcal{B} = [x,y,z]^T$ is mapped to egocentric spherical coordinates $(r, \phi, \theta)$, denoting range $r$, azimuth $\phi$, and elevation $\theta$.
The resulting point cloud is binned into a $2D$ spherical range image $r(i_{\phi}, i_{\theta})$. When multiple points fall into the same bin the one with the smallest distance is chosen. To combine the sensor readings of multiple depth cameras the covered patches corresponding to the individual cameras are added in the range image. For overlapping regions, the pixels with the smallest distance are chosen again. This process is illustrated in the upper part of \cref{fig:architecture}. In practice, to prioritize objects near by, the pixels of the spherical range image are inverted $1/r$ (where $1/r = 0$ corresponds to infinite distance or free space) before feeding it to the policy. 

\subsubsection{Validity Mask}
To explicitly inform the policy of the regions actively covered by the sensors, we construct a binary validity mask $m(i_{\phi}, i_{\theta}) \in \{0, 1\}$ over the canonicalized spherical range image. A mask value of $1$ indicates that a depth measurement was projected into the corresponding angular bin, while unobserved regions are assigned a value of $0$. For a static camera configuration, this validity mask remains constant throughout the evaluation run.


\subsection{Architecture}

\begin{figure}
    \centering
    \includegraphics[width=0.99\linewidth]{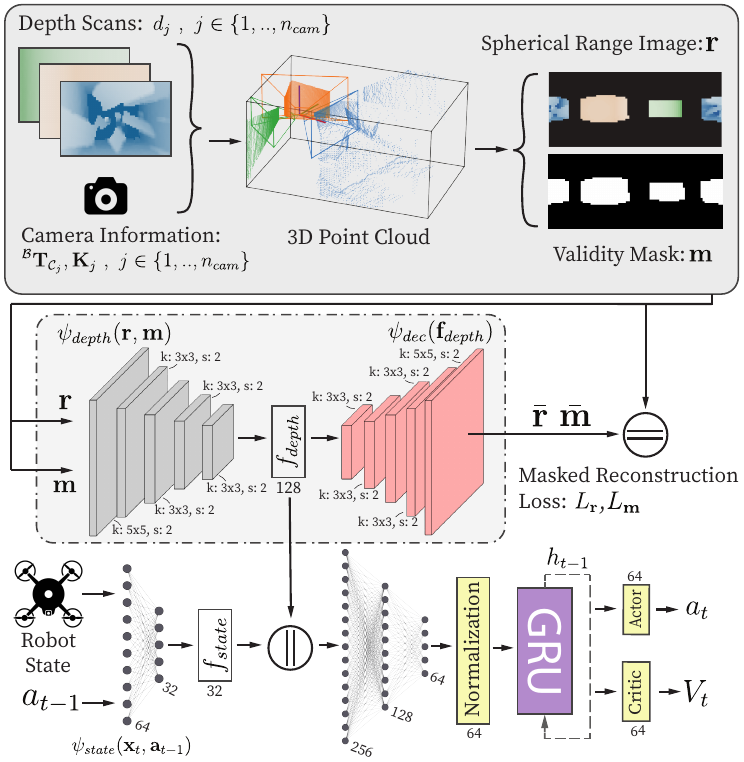}
    \caption{Model architecture.}
    \label{fig:architecture}
\end{figure}

The canonicalized spherical range image $\boldsymbol{r} \in \mathbb{R}^{90\times45}$ and the validity mask $\boldsymbol{m} \in \mathbb{R}^{90\times45}$ are encoded into an image feature directly relating sensor coverage with the full spherical range observation
\vspace{-0.1cm}
\begin{gather}
    \boldsymbol{f}_{depth}=\psi_{depth}(\boldsymbol{r},\boldsymbol{m}) \in \mathbb{R}^{128},
\end{gather}
using the encoder $\psi_{depth}$ shown in \cref{fig:architecture}. To preserve local geometry while maintaining computational efficiency, the encoder consists of five convolutional layers. In each layer we employ circular padding in azimuth to account for the wrap around at $0^\circ$. The pad size $p$ is calculated from the kernel size $k$ as $p = \lfloor k/2 \rfloor$. After each layer we apply $\operatorname{ReLU}$ and the output of the last layer is flattened and then mapped to $\boldsymbol{f}_{depth}$ through a linear layer with an $\operatorname{ELU}$ activation function.
The state estimate of the robot $\boldsymbol{x}_t$ and the previous action $\boldsymbol{a}_{t-1}$ are encoded into a state feature $\boldsymbol{f}_{state}$ using a state encoder $\psi_{state}$
\vspace{-0.1cm}
\begin{gather}
    \boldsymbol{f}_{state}=\psi_{state}(\boldsymbol{x}_t,\boldsymbol{a}_{t-1}) \in \mathbb{R}^{32}.
\end{gather}
with two layers with hidden layer sizes ($64$, $32$). The state and depth features $\boldsymbol{f}_{state}$ and $\boldsymbol{f}_{depth}$ are concatenated and passed through a three-layer feature fusion network with hidden layer sizes ($256$, $128$, $64$) and $\operatorname{ELU}$ activations after each layer. Additionally, a normalization layer is added after the fusion block. Subsequently, the normalized fused features are fed into a $\operatorname{GRU}$ with $64$ hidden units, which outputs the recurrent belief state $\boldsymbol{h}_t$. Finally, $\boldsymbol{h}_t$ is passed to separate actor and critic heads consisting of one linear layer with $64$ units each, outputting the action vector $\boldsymbol{a}_t$ at time $t$ and the state-value estimate $V_t$.

\subsection{Training}
\subsubsection{Reward Design}
For each state transition, the agent receives a scalar reward $R_t$ composed of terms that encourage goal progress, collision avoidance, and visibility-aligned motion, while regularizing action magnitudes and rate of change. The first group of terms encourages goal-directed motion. Let $\boldsymbol{p}_{\mathcal{I}}$ denote the current positions expressed in the inertial frame. Two position-based rewards with different spatial scales are used
\vspace{-0.1cm}
\begin{equation}
d_g^2 =
\left\|
    \boldsymbol{p}_{g} -\boldsymbol{p}_{\mathcal{I}}
\right\|_2^2,
\hspace{0.1cm}
r_{g_i} = \exp\left(-\frac{d_g^2}{\sigma_{g_i}^2}\right),
\hspace{0.1cm} 
i\in\{1,2\},
\end{equation}
where $\sigma_{g_i}$ are parameters defining the width of the kernel function. The smaller value of $\sigma_{g_2}$ makes $r_{g_2}$ more sensitive to the distance near the goal and therefore places greater emphasis on accurate goal tracking. Goal progress is further incentivized by aligning inertial linear velocity $\boldsymbol{v}_{\mathcal{I}}$ with the 
target direction
\vspace{-0.1cm}
\begin{equation}
    r_v = \left\langle \boldsymbol{v}_{\mathcal{I}}, \frac{\boldsymbol{p}_{g} - \boldsymbol{p}_{\mathcal{I}}}{\|\boldsymbol{p}_{g} - \boldsymbol{p}_{\mathcal{I}}\|_2 +\epsilon} \right\rangle,
\end{equation}
where $\epsilon>0$ prevents division by zero. To encourage the commanded linear velocity $\boldsymbol{a}_{x,y,z}$ (linear velocity component of action $\boldsymbol{a}_t$) toward observed space (\cref{fig:act_in_dir}), $r_d = \|\boldsymbol{a}_{x,y,z}\|_2 q(\boldsymbol{a}_{x,y,z})$ scales action magnitude by the valid bin fraction
\vspace{-0.1cm}
\begin{equation}
    q(\boldsymbol{a}_{x,y,z}) = \frac{1}{|\mathcal{W}|} \sum_{(i_{\phi},i_{\theta})\in\mathcal{W}(\boldsymbol{a}_{x,y,z})} m(i_{\phi},i_{\theta}),
\end{equation}
where $m(i_{\phi},i_{\theta}) \in \{0,1\}$ is the validity mask, and $\mathcal{W}(\boldsymbol{a}_{x,y,z})$ is an $n_{cb} \times n_{cb}$ angular window of cardinality $|\mathcal{W}|$ centered on the commanded velocity direction. To penalize motion toward nearby obstacles, a \ac{ttc} penalty $r_c$ is computed across the range grid. The closing speed along each bin's unit ray direction $\boldsymbol{\hat{n}}(i_{\phi},i_{\theta})$ is $s(i_{\phi},i_{\theta}) = \boldsymbol{v}_{\mathcal{B}}^{T}\boldsymbol{\hat{n}}(i_{\phi},i_{\theta})$, where $\boldsymbol{v}_{\mathcal{B}}$ is the body-frame velocity. A bin contributes to the \ac{ttc} calculation only if it contains a valid measurement and the robot is moving toward the corresponding direction. Dropping bin indices for brevity, per-bin \ac{ttc} is defined as
\vspace{-0.1cm}
\begin{equation}
    t_c =
    \begin{cases}
        \dfrac{r}{\max(s,\epsilon)}, & \text{if } s > 0 \text{ and } m = 1, \\[0.5em]
        t_{c_{\max}}, & \text{otherwise},
    \end{cases}
\end{equation}
where $r$ is range, $m \in \{0,1\}$ is validity, $\epsilon > 0$ prevents division by small speeds, and $t_{c_{\max}}$ bounds invalid or non-approaching ($s \le 0$) bins. The minimum predicted \ac{ttc} across all bins determines the penalty
\vspace{-0.1cm}
\begin{equation}
    t_{c_{\min}} = \min_{i_{\phi},i_{\theta}} t_c(i_{\phi},i_{\theta}), \quad 
    r_c = -\exp\left(-\beta_c t_{c_{\min}}^2\right),
\end{equation}
where $\beta_c > 0$ is a kernel scaling parameter. The reward further penalizes collisions and undesirable changes in the commanded actions. A collision that terminates an episode incurs the penalty $r_t=-1$ and $r_t=0$ otherwise. To discourage abrupt changes in the commanded action and unnecessarily large commands, the action-rate and action-magnitude penalties are defined as
\vspace{-0.1cm}
\begin{gather}
    r_r
    =
    -\|\boldsymbol{a}_t-\boldsymbol{a}_{t-1}\|_2^2,
    \qquad
    r_m
    =
    -\|\boldsymbol{a}_t\|_2^2.
\end{gather}
The individual reward terms are combined into the final scalar reward according to
\begin{equation}
\begin{aligned}
R(\boldsymbol{s}_t,\boldsymbol{a}_t) &=
\alpha_{\mathrm{curr}}
\left(
    \alpha_{g_1}r_{g_1} + \alpha_{g_2}r_{g_2} + \alpha_v r_v
\right)\\
&\quad
+\alpha_d r_d + \alpha_c r_c + \alpha_t r_t\\
&\quad
+\alpha_r r_r + \alpha_m r_m,
\end{aligned}
\end{equation}
where $\alpha_{\square}$ denotes tunable weights. In addition to the reward design, an auxiliary reconstruction objective is used during training to encourage the depth encoder to retain information about the observed regions. A decoder block $\psi_{\mathrm{dec}}$ reconstructs the canonicalized range image and validity mask from the latent depth feature $f_{\mathrm{depth}}$. The range reconstruction loss is defined as
\vspace{-0.1cm}
\begin{gather}
    L_{\boldsymbol{r}} = \frac{
        \left\|
            \boldsymbol{m}
            \odot
            \left(
                \bar{\boldsymbol{r}}-\boldsymbol{r}
            \right)
        \right\|_F^2 }{\|\boldsymbol{m}\|_1},
\end{gather}
where $\bar{\boldsymbol{r}}$ denotes the reconstructed range image, $\|\cdot\|_F$ denotes the Frobenius norm, and $\odot$ denotes element-wise multiplication. The validity mask restricts the range reconstruction loss to bins containing valid measurements. The reconstruction of the validity mask is guided via a binary cross-entropy loss
\vspace{-0.1cm}
\begin{equation}
\begin{aligned}
L_{\boldsymbol{m}}&=-\frac{1}{N}
\sum\Big(\boldsymbol{m}\odot\log(\bar{\boldsymbol{m}})\\
&\quad+
    (\boldsymbol{1}-\boldsymbol{m})
    \odot
    \log(\boldsymbol{1}-\bar{\boldsymbol{m}})
\Big),
\end{aligned}
\end{equation}
where $\bar{\boldsymbol{m}}$ denotes the predicted probability of validity and $N$ is the total number of bins in the validity mask. The total loss used to optimize the network weights is then given by
\vspace{-0.1cm}
\begin{gather}
    L=L_{\mathrm{PPO}} + \alpha_L
    \left(
        L_{\boldsymbol{r}}
        +
        L_{\boldsymbol{m}}
    \right),
\end{gather}
where $L_{\mathrm{PPO}}$ denotes the \ac{ppo} loss used to optimize the policy and value networks, and $\alpha_L$ determines the relative contribution of the reconstruction losses.

\subsubsection{Environment Setup}
\begin{figure}
    \centering
    \begin{minipage}[t]{0.25\linewidth}
        \centering
        \includegraphics[width=\linewidth]{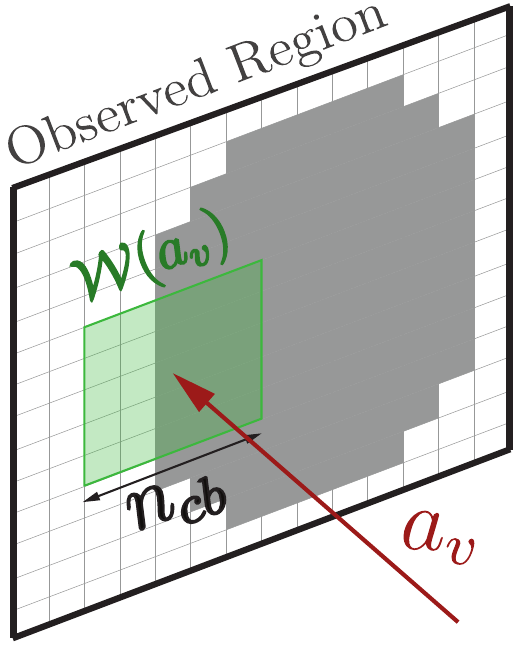}
        \captionof{figure}{Actions in observed region.}
        \label{fig:act_in_dir}
    \end{minipage}
    \hspace{0.5cm}
    \begin{minipage}[t]{0.55\linewidth}
        \centering
        \includegraphics[width=\linewidth]{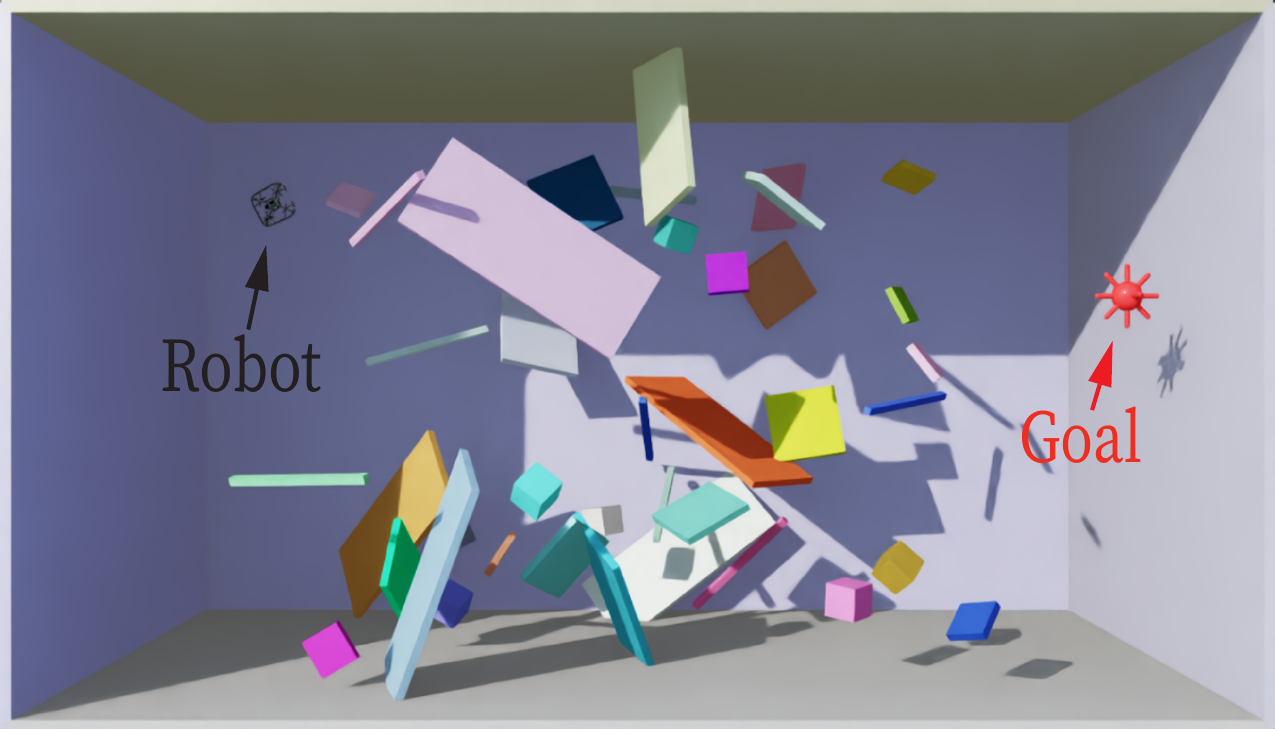}
        \captionof{figure}{Example of a randomly generated obstacle-filled environment.}
        \label{fig:scene_example}
    \end{minipage}
\end{figure}

The training environment consists of a closed room populated with obstacles shown in \cref{fig:scene_example}. The obstacles are placed randomly within the room, with their positions and orientations randomized upon each environment reset. A diverse set of obstacle geometries and sizes is used, ranging from small, thin objects to large, flat panels. The number of obstacles is controlled using a performance-based curriculum: the obstacle density is increased when the policy is able to reliably reach the goal at the current difficulty level. Upon reset a random number of depth cameras is uniformly sampled within the bounds described in \cref{sec:system_representation}. The initial position, orientation and velocities of the robot, as well as the goal position, are randomized at the beginning of each episode. To facilitate sim-to-real transfer, a stereo-depth noise model similar to that of \cite{yang2025spatiallyenhancedrecurrentmemorylongrange} is applied to the depth images generated during training. Moreover, the parameters of the low-level geometric velocity controller are randomized at each environment reset to improve robustness to variations in the system dynamics encountered in the real-world.

\section{Evaluation}
\label{sec:experiments}
To evaluate the proposed generalist navigation policy, we conduct a series of simulation and real-world experiments alongside ablation studies. The policy is trained using our described methodology on a workstation equipped with a single NVIDIA RTX PRO 6000 Blackwell GPU and an Intel Xeon w7-3565X CPU for a total of $1.9 \times 10^6$ environment steps. We utilize Isaac Lab \cite{nvidia2025isaaclabgpuacceleratedsimulation} for GPU-accelerated physics simulation and environment management, which interfaces with the SKRL library \cite{serrano2023skrl} to leverage its implementation of \ac{ppo}. To evaluate a policy we run $1000$ episodes, resampling both the system configuration and the obstacle layout upon every reset. Throughout these trials, we maintain the simulation curriculum at maximum difficulty, resulting in environments consistently populated by $37$ randomly placed obstacles. We report \ac{cr} and \ac{sr}. An episode is considered crashed if a collision occurs and considered successful if it reaches the goal position within $1\mathrm{m}$ radius. Episodes not crashing or succeeding are considered timed out.

\subsection{Ablation Studies}
\begin{figure}[t]
    \centering
    \begin{minipage}[c]{0.4\linewidth}
        \centering
        \vspace*{\fill}
        \scriptsize
        \captionof{table}{Ablation studies. \ac{cr} and \ac{sr} are the mean across all camera numbers ($1,2,3,4$).}
        \label{tab:eval_sim_ablation}
        \begin{tabular}{|c|c|c|} 
        \hline 
        & \ac{cr} & \ac{sr} \\ 
        \hline 
        \hline 
        Ours & $\boldsymbol{0.08}$ & $\boldsymbol{0.88}$ \\ 
        \hline 
        \makecell{Ours w/o mask \\ \& rec. loss} & $0.12$ & $0.76$ \\ 
        \hline 
        Ours w/o rec. loss & $0.10$ & $0.66$ \\ 
        \hline 
        \end{tabular}
        \vspace*{\fill}
    \end{minipage}
    \hfill
    \begin{minipage}[c]{0.45\linewidth}
        \centering
        \vspace*{\fill}
        
        \includegraphics[width=0.9\linewidth]{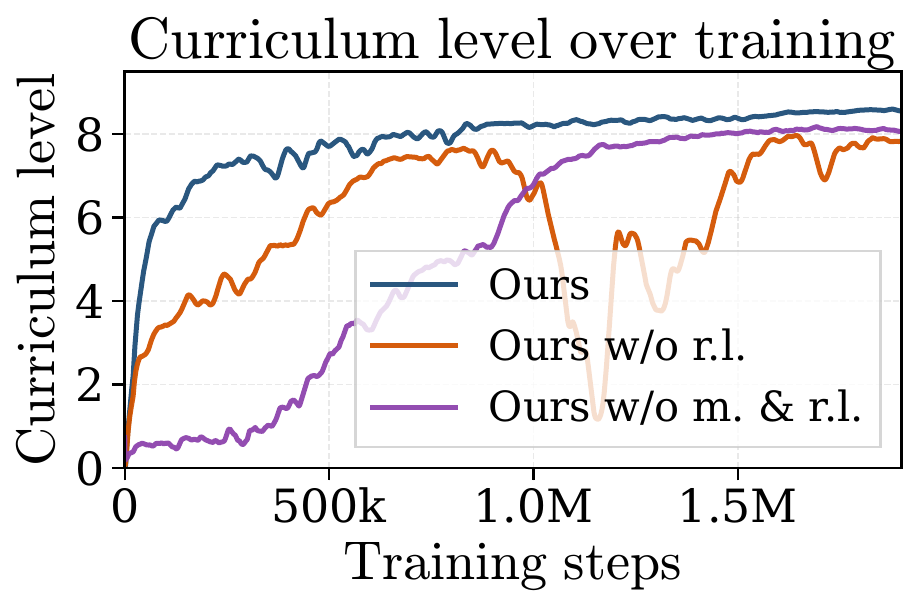}
        \captionof{figure}{Evolution of curriculum levels during training for ablated methods.}
        \label{fig:ablation_plot}
        
        \vspace*{\fill}
    \end{minipage}
\end{figure}

We conduct ablation studies to evaluate the individual contributions of our two primary architectural design choices: the binary validity mask and the auxiliary masked reconstruction loss. Specifically, we evaluate two baseline versions of our method: (i) completely omitting the validity mask from the policy input and also the masked reconstruction loss (denoted as \textit{w/o mask \& rec. loss}). Specifically this means $\boldsymbol{m}$, $L_{\boldsymbol{r}}$ and $L_{\boldsymbol{m}}$ are removed. (ii) providing the validity mask to the network but disabling the guiding auxiliary masked reconstruction loss (denoted as \textit{w/o rec. loss}). Specifically this means $L_{\boldsymbol{r}}$ and $L_{\boldsymbol{m}}$ are removed. Note that in both cases $\boldsymbol{m}$ is still used in the reward calculation. The quantitative results summarized in \cref{tab:eval_sim_ablation} demonstrate that both components are critical to the policy's performance, providing complementary contributions that reduce the \ac{cr} and maximize the overall \ac{sr} over the full training population. \cref{fig:ablation_plot} shows that adding these two components significantly increases sample efficiency and stability of the training process.

\subsection{Simulation Results}
\label{sec:sim_exp}
We evaluate our generalist navigation policy in simulation across the entire family of system configurations considered in this work.

\subsubsection{Performance Evaluation}
To isolate the influence of sensor redundancy, we evaluate the performance for each camera count separately. Notably, we also test the policy on configurations with $5$, $6$, and $7$ cameras; since these layouts are entirely unseen during training, they serve as a rigorous evaluation of the policy's zero-shot generalization to \ac{ood} sensor numbers. \cref{tab:eval_sim_measure} presents the results of this evaluation. As the number of cameras increases, the \ac{cr} decreases and the \ac{sr} increases until performance eventually plateaus. We hypothesize that this trend is driven by the expansion of the combined \ac{fov}; as more cameras are integrated, the aggregate angular coverage (number of bins covered by a sensor divided by total number of bins) increases until additional sensors no longer provide the policy with novel or actionable information about the surrounding environment.
\begin{table}[t]
    \caption{Evaluation of the generalist policy.}
    \centering
    \scriptsize
    \setlength{\tabcolsep}{3pt} 
    \begin{tabular}{|c|c|c|c|c|c|c|c|}
        \hline
        & \multicolumn{7}{c|}{Number of Cameras} \\
        \cline{2-8}
        & $1$ & $2$ & $3$ & $4$ & $5$ (OOD) & $6$ (OOD) & $7$ (OOD) \\
        \hline
        \hline
        \ac{cr}
        & $0.13$ & $0.09$ & $0.06$ & $0.04$ & $0.03$ & $0.04$ & $0.03$ \\
        \hline
        \ac{sr}
        & $0.78$ & $0.87$ & $0.91$ & $0.94$ & $0.95$ & $0.94$ & $0.94$ \\
        \hline
        Coverage 
        & $0.12$ & $0.22$ & $0.31$ & $0.38$ & $0.46$ & $0.50$ & $0.55$\\
        \hline
    \end{tabular}
    \normalsize
    \label{tab:eval_sim_measure}
\end{table}

\subsubsection{Comparison to Specialist Policies}
We further compare the performance of our generalist policy against two classes of specialized baselines:

\noindent\textbf{Comparison with Single-Configuration Specialists:}
First, we compare our generalist policy against configuration-specific specialists trained using our identical methodology, but restricted to a single, fixed camera setup throughout training (i.e., without camera randomization on reset). We sample two random configurations for each camera count $(1,\dots, 4)$, yielding eight distinct configurations for evaluation. The t-SNE visualization of the camera-configuration parameter space is shown as triangles ($\blacktriangle$) in the t-SNE visualization (\cref{fig:t-SNE}). All specialist policies are trained until convergence. The quantitative results, averaged across all eight configurations, are presented in \cref{tab:eval_generalist_ours}. Our generalist policy exhibits a performance penalty of only a $2\%$ increase in \ac{cr} and a $2\%$ decrease in \ac{sr} compared to the specialists, despite being trained to generalize across the entire system family.

\noindent\textbf{Comparison with State-of-the-Art Single-Camera Specialist:}
Second, we compare our method against a \ac{sota} reinforcement learning architecture tailored specifically for single-camera quadrotor navigation \cite{kulkarni2024reinforcement}. To build a comprehensive evaluation set, we sample four additional single-camera configurations, which, combined with the two configurations from our previous evaluation, result in six single-camera test setups. This evaluation set explicitly includes the original, nominal camera configuration used in \cite{kulkarni2024reinforcement} ($0^\circ$ pitch and forward facing camera). These configurations are represented as circles ($\blackcircle$) in the t-SNE embedding (\cref{fig:t-SNE}). We selected the baseline architecture from \cite{kulkarni2024reinforcement} because it was developed and demonstrated in a similar obstacle layout. To ensure a fair and rigorous comparison in simulation, we replace their depth encoder, which was partially pre-trained on real-world images, with a standard convolutional neural network trained entirely end-to-end alongside the policy in simulation. Empirically, this modification yielded superior simulation performance compared to using the original pre-trained weights. The specialist policies are trained until convergence again. The results averaged over all six configurations are summarized in \cref{tab:eval_generalist_kulkarni}. The evaluation demonstrates that our generalist policy performs on par with this highly specialized architecture. Additionally, we compare our approach against the reported \ac{cr} and \ac{sr} in \cite{jacquetNMPC2026} (Table~7, Line~24), which evaluates a system with a forward-facing camera with $0^\circ$ pitch, a wide \ac{fov} ($130^\circ$) for a reference velocity of $1\,\mathrm{m/s}$ in an environment closely matching ours. While both methods achieve a comparable \ac{sr}, their reported \ac{cr} is $6\%$ lower. This comparison is approximate because the reported results in \cite{jacquetNMPC2026} were obtained under a different implementation/environment. This reduced crash rate is likely attributable to their explicit \ac{fov} constraints, which constrain the commanded velocity vector to lie within the camera frustum.
\begin{table}[t]
    \centering
    \scriptsize

    \begin{minipage}[t]{0.48\linewidth}
        \centering
        \caption{Evaluation of our generalist against specialist policies (our method).}
        \begin{tabular}{|c|c|c|}
            \hline
            $\blacktriangle$ & Generalist & Specialist \\
            \hline
            \hline
            \ac{cr} & $0.05$ & $\boldsymbol{0.03}$ \\
            \hline
            \ac{sr} & $0.93$ & $\boldsymbol{0.95}$ \\
            \hline
        \end{tabular}
        \label{tab:eval_generalist_ours}
    \end{minipage}
    \hfill
    \begin{minipage}[t]{0.48\linewidth}
        \centering
        \caption{Evaluation of our generalist against specialist policies trained with a modified version of the architecture in \cite{kulkarni2024reinforcement}.}
        \begin{tabular}{|c|c|c|}
            \hline
            \blackcircle & Generalist & Specialist \\
            \hline
            \hline
            \ac{cr} & $\boldsymbol{0.08}$ & $\boldsymbol{0.08}$ \\
            \hline
            \ac{sr} & $\boldsymbol{0.9}$ & $0.88$ \\
            \hline
        \end{tabular}
        \label{tab:eval_generalist_kulkarni}
    \end{minipage}
\end{table}

\subsubsection{Design Space Evaluation}
\begin{figure}
    \centering
    \includegraphics[width=\linewidth]{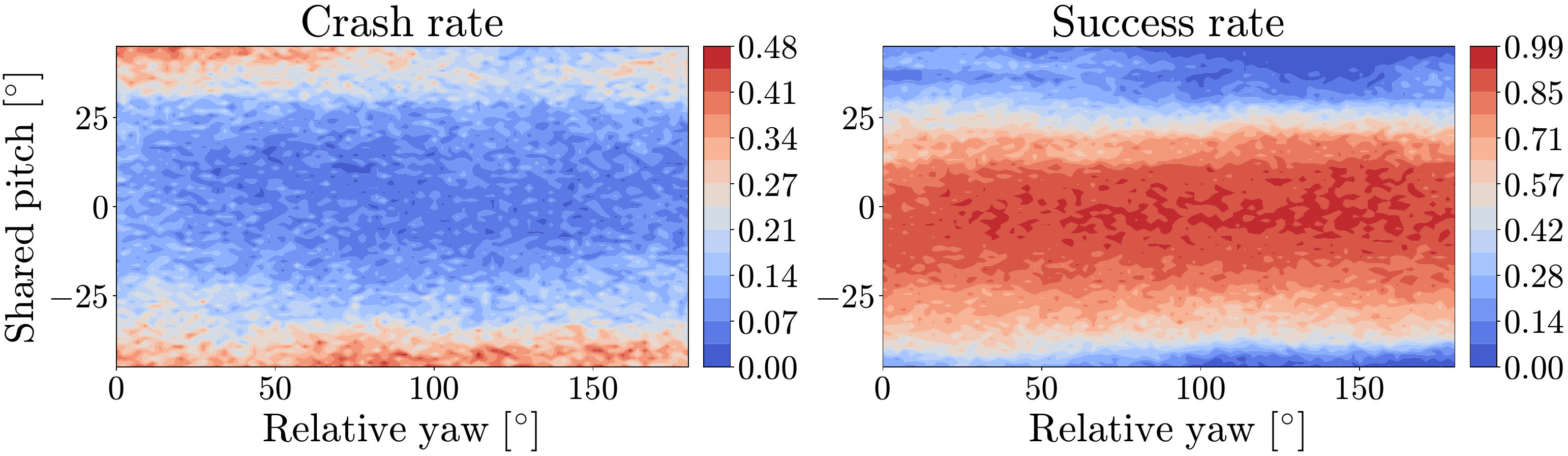}
    \caption{Reported \ac{cr} and \ac{sr} for the design space evaluation.}
    \label{fig:design_space_eval}
\end{figure}

To visualize policy performance across a dense design space, we evaluate two-camera configurations sweeping shared pitch and relative yaw angles ($100$ closed-loop episodes per setup; see \cref{fig:design_space_eval}). As camera pitch diverges from the body's $x-y$ plane, \ac{cr} increases and \ac{sr} decreases. This stems from underactuated quadrotor dynamics, where attitude pitch is rigidly coupled to horizontal acceleration: non-zero mounting offsets force continuous attitude adjustments to align view axes with the velocity vector. Performance also degrades at extreme relative yaw angles due to either redundant \ac{fov} overlap (narrow yaw) or loss of forward-flight coverage (at extreme wide yaw angles, the cameras point in opposite directions).

\subsubsection{Sensitivity Analysis}
\begin{figure}
    \centering
    \includegraphics[width=0.99\linewidth]{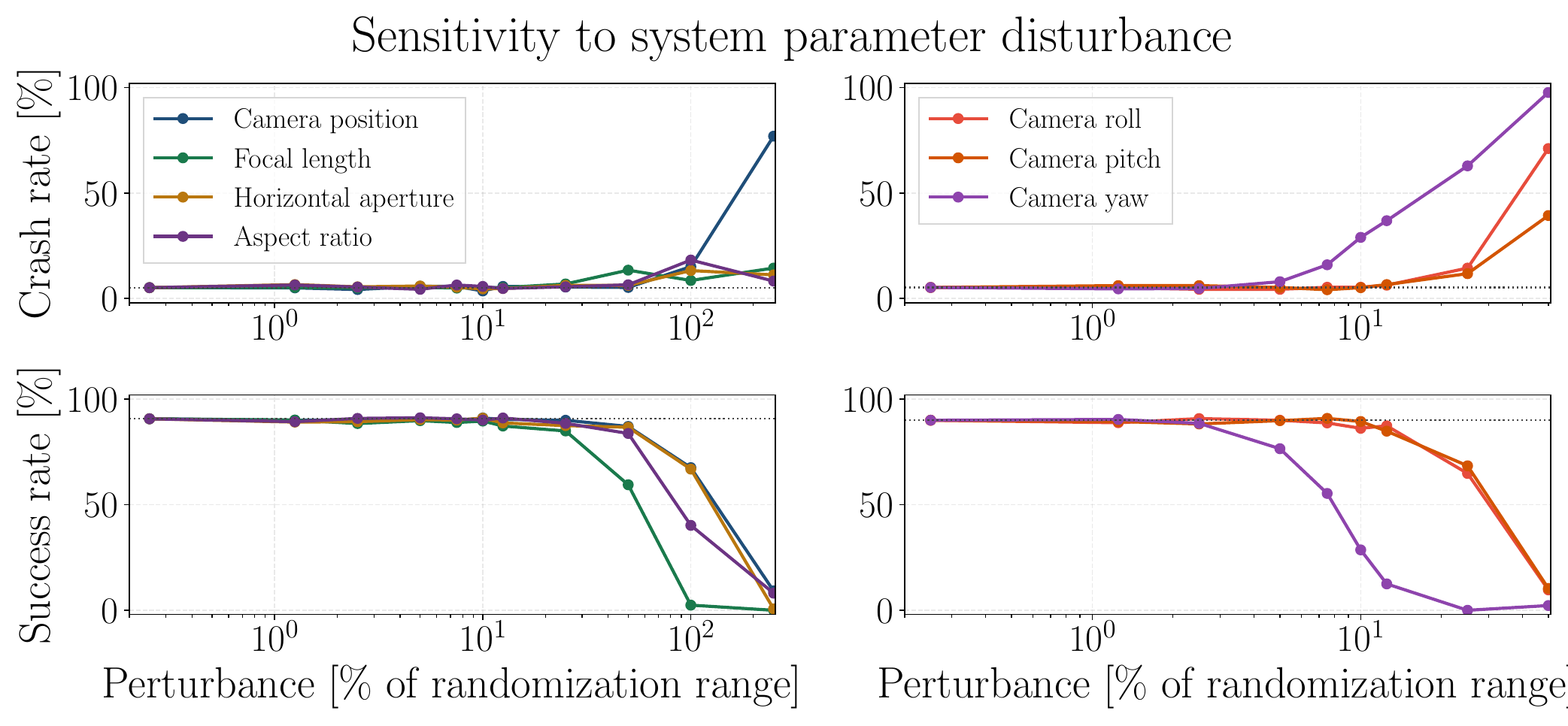}
    \caption{Analysis of the sensitivity to parameter disturbances.}
    \label{fig:sensitivity_analysis}
\end{figure}

To evaluate how mismatches between the estimated embodiment parameters and the platform's true physical parameters affect policy performance, we conduct a sensitivity analysis. Specifically, we track the evolution of the \ac{cr} and \ac{sr} as perturbations of increasing magnitude are applied to the system parameters used to compute the canonicalized range image and the validity mask. Perturbation magnitudes are defined as a percentage of each parameter's randomization range (see \cref{tab:camera_randomization}), and the sign of each disturbance is sampled with equal probability. The resulting \ac{cr} and \ac{sr} curves are shown in \cref{fig:sensitivity_analysis}. The analysis indicates no performance drop for parameter mismatches of up to $10\mathrm{\%}$ of the randomization range for all intrinsic parameters and for the camera position. For camera orientation, the method remains robust to yaw mismatches of up to $2.5\mathrm{\%}$ and to roll and pitch mismatches of up to $7.5\mathrm{\%}$ of their respective randomization ranges. These results highlight the robustness of the proposed approach and support its suitability for sim-to-real transfer. 
\begin{figure}
    \centering
    \begin{minipage}[t]{0.49\linewidth}
        \centering
        \includegraphics[width=\linewidth]{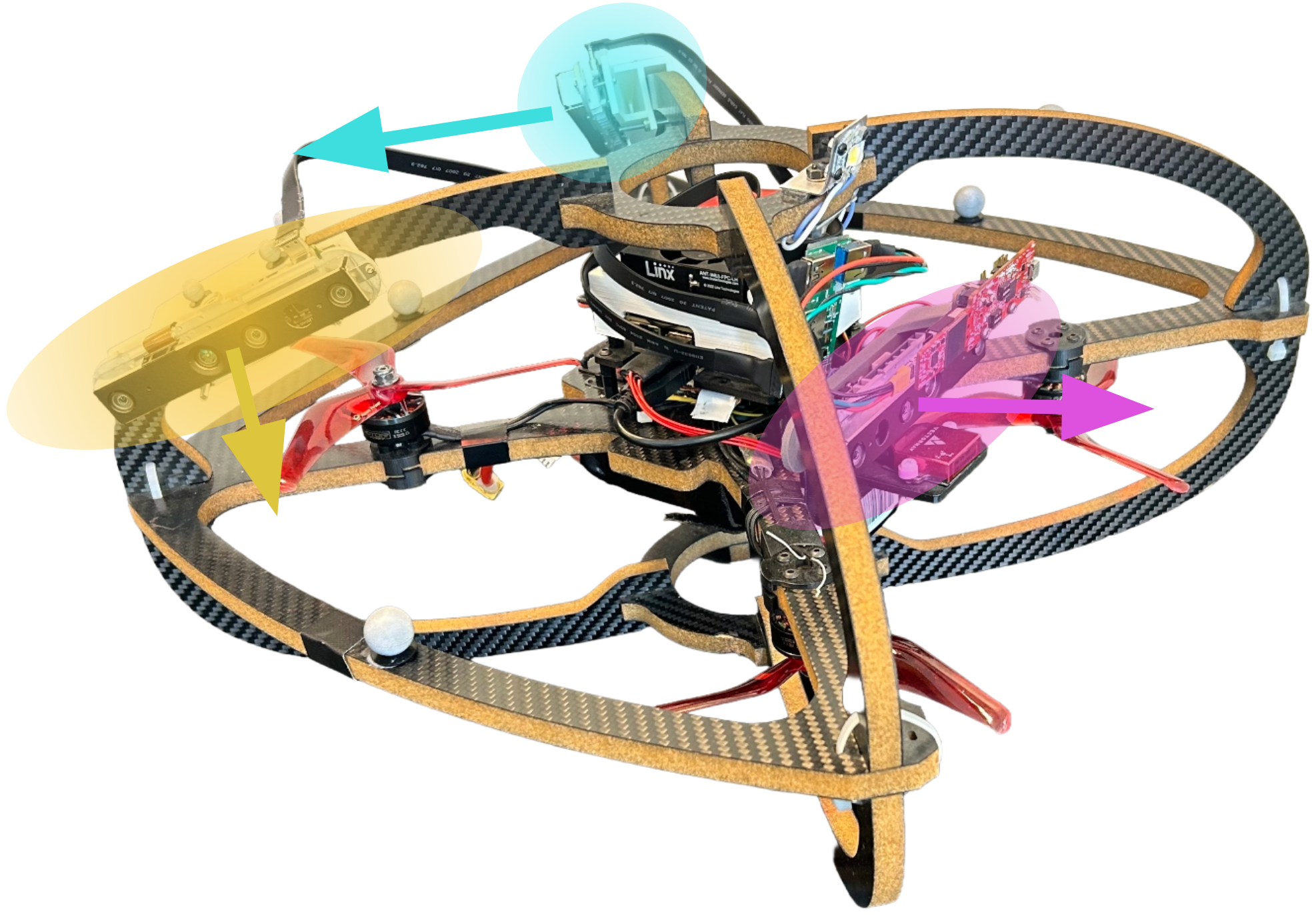}
        \caption{Real system used for deployment. Camera $1$: magenta, camera $2$: yellow, camera $3$: blue.}
        \label{fig:real_system}
    \end{minipage}
    \hfill
    \begin{minipage}[t]{0.49\linewidth}
        \centering
        \includegraphics[width=\linewidth]{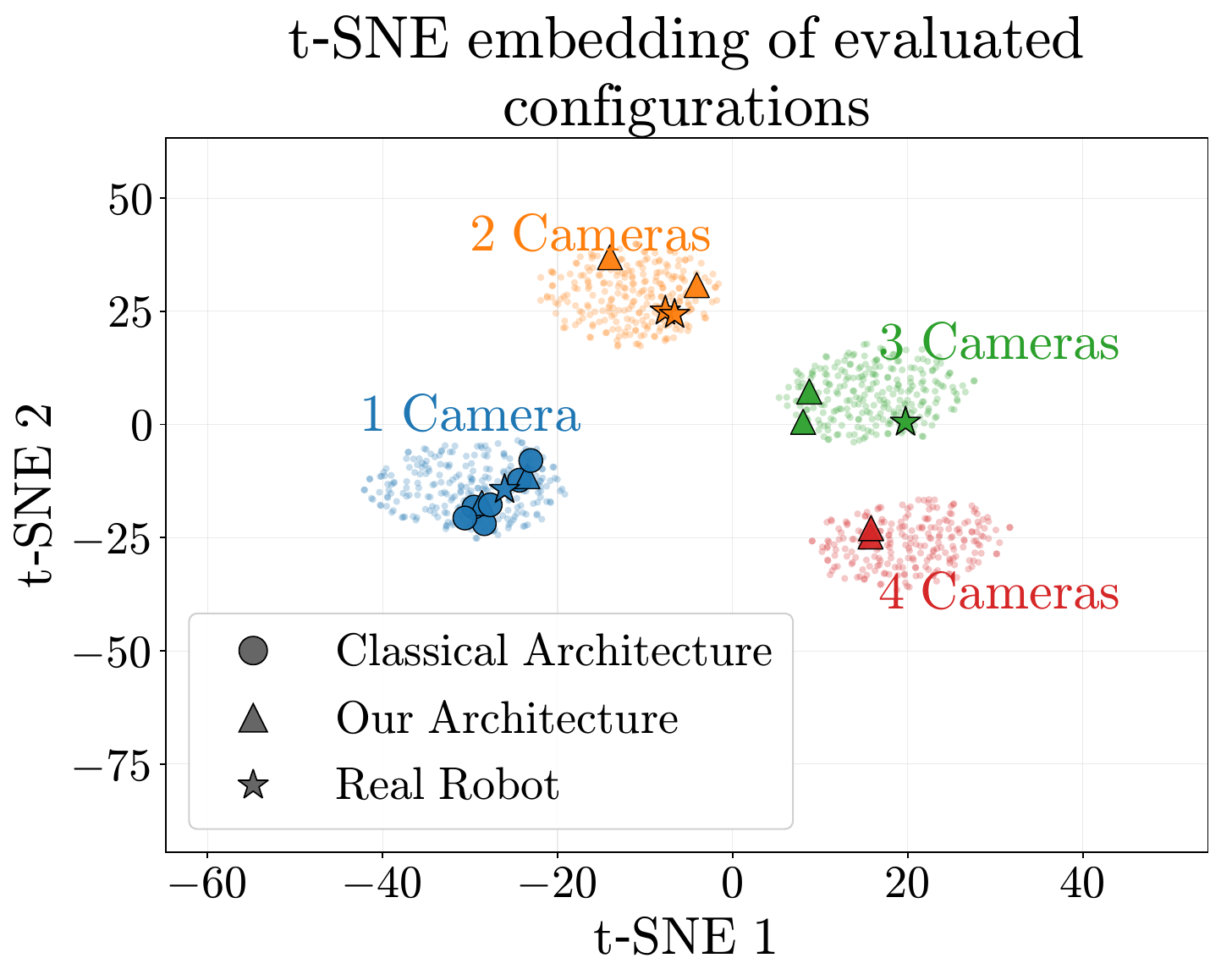}
        \captionof{figure}{t-SNE embedding of evaluated configurations. Note that the parameter vector $\gamma$ is zero padded to unify different camera numbers.}
        \label{fig:t-SNE}
    \end{minipage}
\end{figure}

\subsection{Real-World Experiments}
\label{sec:real_exp}
To validate the real-world performance of our generalized navigation policy, we first deploy it on a custom quadrotor platform equipped with three Intel RealSense D455 depth cameras, an NVIDIA Jetson Orin NX onboard computer, a Texas Instruments IWR6843AOP-EVM radar, and a Pixracer Pro flight controller running the PX4 autopilot firmware. State estimation is resolved onboard using \ac{rio}, and low-level velocity tracking commands are executed directly by the PX4 controller. Our policy runs entirely onboard, performing inference at $10~\text{Hz}$. The three cameras are physically arranged as illustrated in \cref{fig:real_system}. By selectively disabling individual camera streams, we construct three distinct hardware configurations to evaluate zero-shot adaptation: (i) all three cameras active, (ii) cameras $1$ and $2$ active, and (iii) only camera $3$ active. The t-SNE visualization of the camera-configuration parameter space is visualized in the t-SNE embedding in \cref{fig:t-SNE} (indicated by the stars $\blackstar$). We evaluate the system across two highly distinct real-world environments: a cluttered indoor corridor and a dense forest. 

\subsubsection{Cluttered Hallway}
First, we evaluate the generalist policy across all three sensor configurations in an indoor hallway cluttered with obstacle panels, with vertical ($z$) actions constrained to prevent flying over obstacles (\cref{fig:all_combined} a)). When initialized with its single active camera pointing away from the goal, the policy demonstrates safety-first behavior: rather than translating blindly into unobserved space, it executes an in-place yaw rotation to align its camera with the target before initiating forward movement. In all environments, the policy smoothly avoids the obstacles in its way.
\textbf{Robustness to Sensor Dropout:} 
Additionally, we evaluate policy robustness under online sensor dropout. For this we first re-run the $2$-, $3$-, and $4$-camera experiments in simulation while randomly disabling one active sensor mid-episode (removing the depth readings from the canonicalized observations and setting the corresponding mask pixels to $0$). Despite the sudden loss of sensor coverage, the policy adapts online in most of the cases: $2$-camera setups achieve a \ac{cr} of $0.15$ and \ac{sr} $0.74$, $3$-camera setups reach a \ac{cr} of $0.10$ and \ac{sr} $0.85$, and $4$-camera setups achieve a \ac{cr} of $0.05$ and \ac{sr} $0.92$. We also demonstrate dropout robustness in two real-world scenarios (\cref{fig:all_combined} b)). In the first, camera 3 abruptly fails mid-trajectory on a three-camera setup. In the second, camera $1$ is disabled mid-flight while camera $3$, pointing in the opposite direction, is simultaneously activated, instantaneously reversing the sensing envelope. The policy handles this shift safely: it immediately halts forward translation, executes a $\sim 180^\circ$ yaw rotation to align the newly active camera with the target, and resumes navigation. 
\textbf{Generalization to Different Depth Sensors:} To demonstrate our methods capacity to generalize across heterogeneous depth-sensing hardware in reality, we replace camera $2$ (Intel RealSense D455) with a pmd flexx2 \ac{tof} sensor at the identical physical mounting position. This \ac{tof} sensor features a significantly narrower \ac{fov} ($56^\circ \times 44^\circ$) compared to the D455 ($87^\circ \times 58^\circ$). We conduct an experiment utilizing a heterogeneous sensor setup consisting of the newly integrated pmd flexx2 and camera $1$ (D455) (t-SNE embedding shown in \cref{fig:t-SNE}). The successful navigation of the platform shown in \cref{fig:all_combined} d) demonstrates that our canonicalization pipeline effectively adapts to varying sensor models, physical resolutions, and fields of view without requiring retraining.

\subsubsection{Forest}
\begin{figure*}
    \centering
    \includegraphics[width=0.95\linewidth]{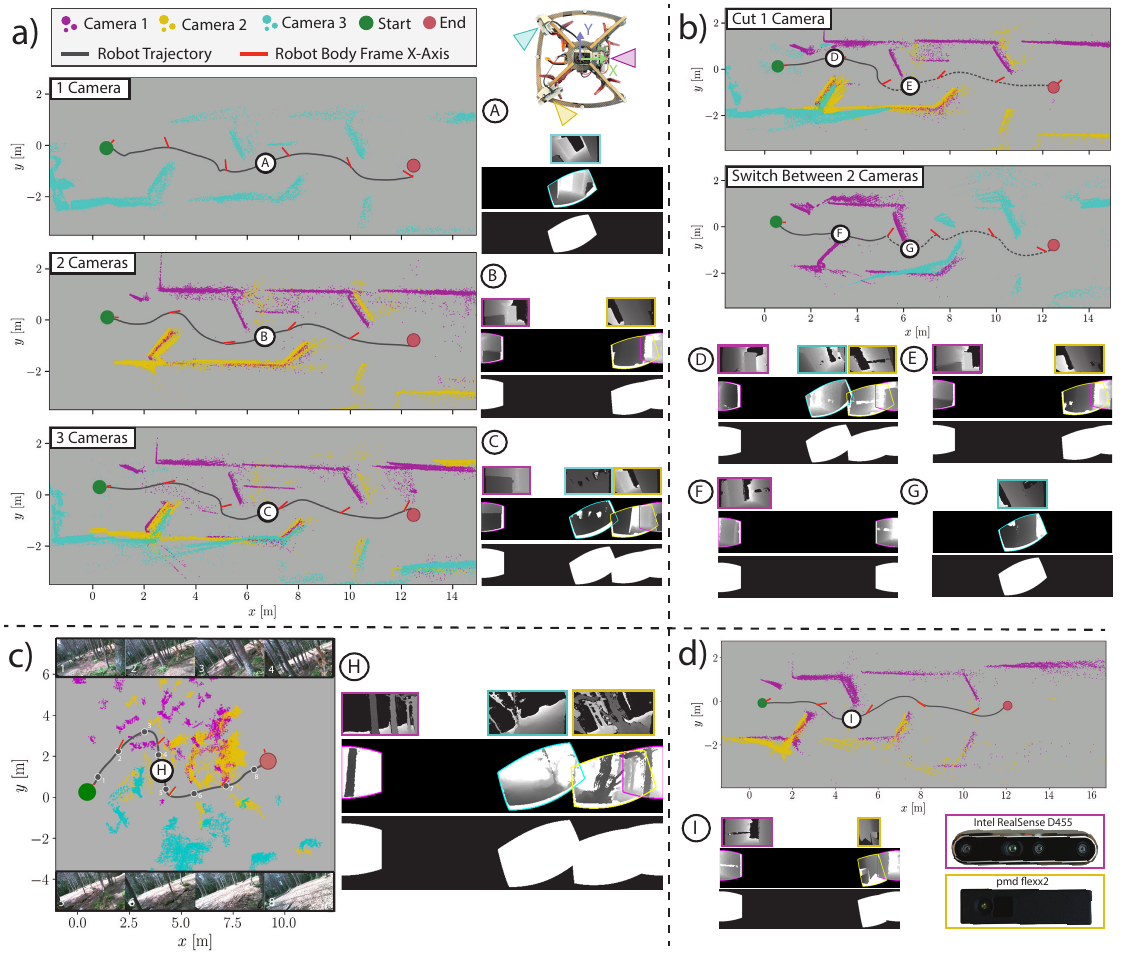}
    \caption{Real-world experiments: a) cluttered hallway, b) sensor dropout, c) forest, d) heterogeneous sensor config using an Intel RealSense D455 and a pmd flexx2. Range images and validity masks are cropped vertically and shown at native resolution for visual clarity. The RGB images for the forest result are taken with camera $2$.}
    \label{fig:all_combined}
\end{figure*}

To evaluate the policy's zero-shot generalization to \ac{ood} geometric structures, we deployed our system in a dense natural forest. Note that we remove the constraint on the $z$-velocity in this environment. Unlike the primitive, geometric obstacles (e.g., cylinders and planar barriers) encountered during simulation training, natural forest environments present highly complex, non-convex topologies characterized by thin tree trunks, overhanging branches, and irregular foliage. Operating entirely without fine-tuning, our generalist policy successfully navigates this challenging environment. As illustrated in \cref{fig:all_combined} c), the policy leverages multiple sensor streams to navigate around unobserved spatial regions and thin obstacles, executing smooth, continuous 3D avoidance maneuvers while maintaining target forward progress.

\section{Conclusion and Future Work}
\label{sec:conclusion}

In this work, we presented a camera-layout-agnostic policy for reactive local collision avoidance across heterogeneous depth sensor configurations. By unprojecting depth measurements into a canonical egocentric spherical range image paired with a binary validity mask, our approach eliminates the need to implicitly learn camera alignment, enabling zero-shot adaptation to varying camera counts, extrinsics, and intrinsics. Simulation evaluations confirmed zero-shot generalization to \ac{ood} sensor payloads with up to seven cameras. Physical quadrotor flight trials in cluttered indoor corridors and dense forests demonstrated robustness to complex geometries and resilience to sudden online camera dropouts. Future work will couple this exteroceptive canonicalization with morphology-aware dynamics to achieve joint transfer across both sensor layouts and physical robot platforms. 

\bibliographystyle{IEEEtran}
\bibliography{bibliography/Library}

@misc{nvidia2025isaaclabgpuacceleratedsimulation,
      title={Isaac Lab: A GPU-Accelerated Simulation Framework for Multi-Modal Robot Learning}, 
      author={{NVIDIA} and Mayank Mittal and Pascal Roth and James Tigue and Antoine Richard and others},
      year={2025},
      eprint={2511.04831},
      archivePrefix={arXiv},
}

@article{yang2025spatiallyenhancedrecurrentmemorylongrange,
  author = {Yang, Fan and Frivik, Per and Hoeller, David and Wang, Chen and Cadena, Cesar and Hutter, Marco},
  title = {Spatially-enhanced recurrent memory for long-range mapless navigation via end-to-end reinforcement learning},
  journal = {The International Journal of Robotics Research},
  year = {2025},
}

@INPROCEEDINGS{ShahGNM2023,
  author={Shah, Dhruv and Sridhar, Ajay and Bhorkar, Arjun and Hirose, Noriaki and Levine, Sergey},
  booktitle={2023 IEEE International Conference on Robotics and Automation (ICRA)}, 
  title={{GNM}: A General Navigation Model to Drive Any Robot}, 
  year={2023},
  volume={},
  number={},
  pages={7226-7233},
  doi={10.1109/ICRA48891.2023.10161227}}

@article{raptor,
  title={RAPTOR: A foundation policy for quadrotor control},
  author={Eschmann, Jonas and Albani, Dario and Loianno, Giuseppe},
  journal={Science Robotics},
  year={2026},
  publisher={American Association for the Advancement of Science}
}

@inproceedings{
bohlinger2025policyrunallendtoend,
title={One Policy to Run Them All: an End-to-end Learning Approach to Multi-Embodiment Locomotion},
author={Nico Bohlinger and Grzegorz Czechmanowski and Maciej Piotr Krupka and Piotr Kicki and Krzysztof Walas and Jan Peters and Davide Tateo},
booktitle={8th Annual Conference on Robot Learning},
year={2024},
}

@inproceedings{luo2025gcntgraphbasedtransformerpolicies,
  author    = {Luo, Yingbo and Yao, Meibao and Xiao, Xueming},
  title     = {GCNT: Graph-Based Transformer Policies for Morphology-Agnostic Reinforcement Learning},
  booktitle = {Proceedings of the Thirty-Fourth International Joint Conference on Artificial Intelligence},
  year      = {2025},
}

@inproceedings{
eftekhar2025the,
title={The One {RING}: a Robotic Indoor Navigation Generalist},
author={Ainaz Eftekhar and Luca Weihs and Rose Hendrix and Ege Caglar and Jordi Salvador and Alvaro Herrasti and Winson Han and Eli VanderBilt and Aniruddha Kembhavi and Ali Farhadi and Ranjay Krishna and Kiana Ehsani and Kuo-Hao Zeng},
booktitle={The first CVPR workshop on 3D Vision Language Models (VLMs) for Robotics Manipulation: Opportunities and Challenges},
year={2025},
}

@inproceedings{
zhang2026embodied,
title={Embodied Navigation Foundation Model},
author={Jiazhao Zhang and Anqi Li and Yunpeng Qi and Minghan Li and Jiahang Liu and Shaoan Wang and Haoran Liu and Gengze Zhou and Yuze Wu and Xingxing LI and Yuxin Fan and Wenjun Li and Zhibo Chen and Fei Gao and Qi Wu and Zhizheng Zhang and He Wang},
booktitle={The Fourteenth International Conference on Learning Representations},
year={2026},
}

@ARTICLE{Wang2026XNav,
  author={Wang, Haitong and Tan, Aaron Hao and Fung, Angus and Nejat, Goldie},
  journal={IEEE Robotics and Automation Letters}, 
  title={X-Nav: Learning End-to-End Cross-Embodiment Navigation for Mobile Robots}, 
  year={2026},
  volume={11},
  number={1},
  pages={698-705}}

@inproceedings{shah2023vint,
          title     = {Vi{NT}: A Foundation Model for Visual Navigation},
          author    = {Dhruv Shah and Ajay Sridhar and Nitish Dashora and Kyle Stachowicz and Kevin Black and Noriaki Hirose and Sergey Levine},
          booktitle = {7th Annual Conference on Robot Learning},
          year      = {2023},
        }

@article{serrano2023skrl,
  author  = {Antonio Serrano-Muñoz and Dimitrios Chrysostomou and Simon Bøgh and Nestor Arana-Arexolaleiba},
  title   = {skrl: Modular and Flexible Library for Reinforcement Learning},
  journal = {Journal of Machine Learning Research},
  year    = {2023},
  volume  = {24},
  number  = {254},
  pages   = {1--9},
}

@INPROCEEDINGS{kulkarni2024reinforcement,
  author={Kulkarni, Mihir and Alexis, Kostas},
  booktitle={2024 IEEE International Conference on Robotics and Automation (ICRA)}, 
  title={Reinforcement Learning for Collision-free Flight Exploiting Deep Collision Encoding}, 
  year={2024},
  volume={},
  number={},
  pages={15781-15788}}

@misc{konstantaropoulos2026embodiment,
      title={Embodiment-conditioned Generalist Control for Multirotor Aerial Robots}, 
      author={Orestis Konstantaropoulos and Welf Rehberg and Mihir Kulkarni and Kostas Alexis},
      year={2026},
      eprint={2606.10857},
      archivePrefix={arXiv},
      primaryClass={cs.RO},
      url={https://arxiv.org/abs/2606.10857}, 
}

@INPROCEEDINGS{zhang2023learning,
  author={Zhang, Dingqi and Loquercio, Antonio and Wu, Xiangyu and Kumar, Ashish and Malik, Jitendra and Mueller, Mark W.},
  booktitle={2023 IEEE International Conference on Robotics and Automation (ICRA)}, 
  title={Learning a Single Near-hover Position Controller for Vastly Different Quadcopters}, 
  year={2023}}

@article{jacquetNMPC2026,
author = {Martin Jacquet and Marvin Harms and Kostas Alexis},
title ={Neural NMPC through signed distance field encoding for collision avoidance},
journal = {The International Journal of Robotics Research},
volume = {45},
number = {10},
pages = {1597-1623},
year = {2026}
}

@article{
loquercioLearning2021,
author = {Antonio Loquercio  and Elia Kaufmann  and René Ranftl  and Matthias Müller  and Vladlen Koltun  and Davide Scaramuzza },
title = {Learning high-speed flight in the wild},
journal = {Science Robotics},
volume = {6},
number = {59},
year = {2021}
}

@Inbook{florenceIntegrated2020,
author="Florence, Pete
and Carter, John
and Tedrake, Russ",
title="Integrated Perception and Control at High Speed: Evaluating Collision Avoidance Maneuvers Without Maps",
bookTitle="Algorithmic Foundations of Robotics XII: Proceedings of the Twelfth Workshop on the Algorithmic Foundations of Robotics",
year="2020",
publisher="Springer International Publishing",
pages="304--319",
}

@INPROCEEDINGS{hengAutonomous2011,
author={Heng, Lionel and Meier, Lorenz and Tanskanen, Petri and Fraundorfer, Friedrich and Pollefeys, Marc},
booktitle={2011 IEEE International Conference on Robotics and Automation}, 
title={Autonomous obstacle avoidance and maneuvering on a vision-guided MAV using on-board processing}, 
year={2011},
pages={2472-2477}}

\end{document}